# Classical and quantum regression analysis for the optoelectronic performance of NTCDA/p-Si UV photodiode

Ahmed M. El-Mahalawy[1], Kareem H. El-Safty[2,3,4*]

1. Thin Film Laboratory, Physics Department, Faculty of Science, Suez Canal University, Ismailia, Egypt.
2. Wigner Research Centre for Physics, Budapest, Hungary
3. Artificial Intelligence Department, DevisionX, Cairo, Egypt.
4. Alexandria Quantum Computing Group, Faculty of Science, Alexandria University, Alexandria, Egypt.

## Abstract

Due to the pivotal role of UV photodiodes in many technological applications in tandem with the high efficiency achieved by machine learning techniques in regression and classification problems, different artificial intelligence techniques are adopted to simulate and model the performance of organic/inorganic heterojunction UV photodiode. Herein, the performance of a fabricated Au/NTCDA/p-Si/Al photodiode is explained in a detailed manner and has shown an excellent responsivity and detectivity for UV light of intensities; ranging from 20 to 80 $mW/cm^2$ . A linear current–irradiance relationship is exhibited by the fabricated photodiode under illumination up to 65 $mW/cm^2$ . It also shows good response times of $t_{rise} = 408\ ms$ and $t_{fall} = 490\ ms$. Furthermore, we have not only fitted the characteristic *I-V* curve but also evaluated three classical algorithms; K-Nearest Neighbour, Artificial Neural Network, and Genetic Programming besides using a Quantum Neural Network to predict the behaviour of the device. The models have achieved outstanding results and managed to capture the trend of the target values. The Quantum Neural Network has been used for the first time to model the photodiode. The models can be used instead of repeating the fabrication process. This means a reduction in the cost and the manufacturing time.

**Keywords**: Organic Semiconductor – Heterojunction Photodiode – Machine Learning – Genetic Programming – Quantum Machine Learning.

**First Author:**
Name: Ahmed M. El-Mahalawy
Thin Film Laboratory, Physics Department, Faculty of Science, Suez Canal University, Ismailia, Egypt.
Email: ahmed_el.mahalawy@yahoo.com
ORCID: 0000-0002-2613-0116

**Second & Corresponding Author:**
Name: Kareem H. El-Safty.
Wigner Research Centre for Physics, Budapest, Hungary
Email: kareem.elsafty@wigner.hu
Tel: +201123398119
ORCID: 0000-0001-8740-0637

# 1  Introduction

Over the past years, the technology of optoelectronic devices based on planar organic semiconductor has witnessed an escalating pace both in academia and industry. Notably, the fusion of both organic and inorganic materials to form the hybrid heterojunction devices has triggered a stir in the field of photodiode fabrication research because of their capability of tuning the spectral range of detection by changing the deposited organic film on the inorganic substrate [1]. From the multiple choices of planar structured organic materials, naphthalenetetracarboxylic dianhydride (NTCDA) has favourable features. 1,4,5,8-naphthalenetetracarboxylic dianhydride with a planar π-stacking structure has emerged as a promising n-type organic semiconductor material for the organic electronics community.

Recently, there were several studies concerning the explanation of many physical and chemical aspects of this compound such as the degradation mechanism, hybrid complexation with inorganics and properties of monolayers epitaxial grown [2–6]. Furthermore, the unique characteristics of NTCDA such as the high electron mobility [7,8], the high UV absorption [9,10], the high air stability [7,11] and the obvious photocurrent multiplication in optical photodiodes [12,13] recommend this material for a wide range of optoelectronic applications.

In this framework, NTCDA is utilised in various microelectronics and optoelectronic applications such as organic light-emitting diodes (OLEDs) [14], organic photodiodes [15–17], organic solar cells (OSCs) [18,19], organic diodes [20] and n-channel organic field-effect transistor's (OFETs) [11,21–23]. According to the features mentioned above of NTCDA, the present study introduces a hybrid heterojunction based on NTCDA/p-Si for photodetection applications.

On another side, for decades, modelling techniques that depend mainly on fitting criteria have been used in physics. Lately, more sophisticated approaches have been adopted to replace the fitting methods in computational physics. The work mentioned in [24–30] proves that the field of Artificial Intelligence (AI) and especially Machine Learning (ML) achieved high efficiency in modelling and fitting problems in multidiscipline areas. ML is considered a vital tool that transforms the way we treat and synthesise new materials. ML models; either straightforward algorithms or Deep Learning (DL) algorithms do not only fit the data but also can predict the behaviour of microelectronic devices and give us an intuition behind the inner workings of physical phenomenon.

In a retrospective manner, the relationship between condensed matter physics and ML can be exploited by the mutual benefits of using the concepts of each one to boost the progress of the other. Examples for this are the emergence of Quantum Machine Learning (QML) [31–33] to improve the speed of some algorithms and achieve higher accuracies and using ML for discovering new materials or optimising the manufacturing process for such materials.

The present study introduces the utilisation of four different ML algorithms to model the behaviour of a fabricated hybrid heterojunction UV photodiode. Besides that, we analyse the optoelectronic characteristics of the fabricated photodiode. The approach for the modelling part is based on regression analysis. This means that we no longer look for the lowest error rate but for a model that can generalise well enough to predict the behaviour of the photodiode and can be used by others to conduct their own studies without the need of repeating the costly manufacturing process.

The algorithms are divided into two main groups; classical algorithms and a single quantum algorithm. The classical algorithms are K-Nearest Neighbour (KNN) [34,35], Neural Networks (NN) [27,36], and Automated Machine Learning (AML) using Genetic Programming (GP) [37]. The quantum algorithm is The Continuous-Variable (CV) Quantum Neural Network (QNN) [32]. The paper is divided as follows: the upcoming section will be the materials and methods where we will introduce the fabrication process and the software modelling requirements and procedures. In section three we will discuss the characterisation techniques and the algorithms for the modelling part at length. The optoelectronic analysis and the merits of the fabricated photodiode alongside the ML results from the modelling algorithms will be highlighted in section four. Finally, we will summarise our opinions and important results.

## 2 Materials and Methods

The architectural design of the fabricated photodiode is illustrated in Fig. 1.a, where the p-type silicon substrates were carefully etched and cleaned as mentioned in our earlier work [38]. The chemical structure of 1,4,5,8-naphthalene tetracarboxylic-dianhydride, NTCDA, ($C_{14}H_4O_6$) powder purchased from *Alfa Aeser* Company is shown in Fig. 1.b. A thermally evaporated thin film of NTCDA of thickness $150\ nm$ was deposited using the coating unit (model: Edwards 306A-England) at $1 \times 10^{-5}\ mbar$ and room temperature. Then, a $50\ nm$-thick gold top electrode was deposited on NTCDA thin film using a shadow mask in the form of a mesh. Finally, a thick aluminium bottom electrode of thickness $300\ nm$ was deposited on the backside of the Si-substrate.

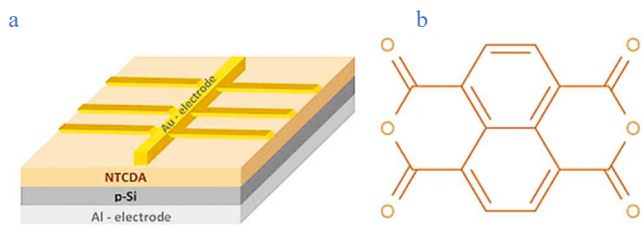

**Fig**. 1: (a) Architecture of the fabricated photodiode, and (b) the molecular structure of NTCDA.

The software language used in this research is Python 3.6. We have used Keras, Scikit-Learn, Pennylane, and Strawberryfields [39–42] to model the characteristics of the photodiode. For verification and replicating the results, the code and the dataset are available on our [GitHub](#) repository.

## 3 Experimental work

### 3.1 Characterisation techniques

The optical absorbance of NTCDA thin film on a quartz substrate was measured at normal incidence of the light at room temperature in the spectral range of $190$– $2500\ nm$ using a double beam spectrophotometer (*JASCO model V-570 UV-VIS-NIR*). The Photoelectrical properties – of the fabricated Au/NTCDA/p-Si/Al photodiode – were investigated by measuring the *I-V* characteristic curve at room temperature from $-3.5\ V$ to $+3.4\ V$ using Keithley electrometer model *6517B* under the influence of UV light of wavelength $194\ nm$. The intensity of incident light was measured using the Radiometer / Photometer model IL1400A.

### 3.2 Modelling procedures

#### 3.2.1 Data inspection

The approach for this section is entirely data-driven and is built on publicly available software. We adopted the standard pipeline for ML which consists of data acquisition, data cleansing, feature Engineering and pre-processing, model selection, hyperparameter tuning, and validation and testing. In our case, the first two steps have been done using Keithley electrometer model *6517B*. To decide what feature pre-processing method should be used, we must conduct an explanatory data analysis (EDA) approach; so that we can summarise the characteristics and the insights of the dataset.

Based on the dataset characteristics that are represented in table 1, one can speculate about the model and its properties and then find a suitable range of parameters to tune the model against a fitness function to have the optimum set of values for it. Finally, the model can be validated and tested against out of sample data.

*Table.* 1 The estimated statistical parameters of the I-V samples.

| $Stats$ | $V\ (Volt)$ | $Illumination\ Intensity\ (mW/cm^2)$ | $I\ (amp)$ |
|---|---|---|---|
| **No. Samples** | 828 | 828 | 828 |
| $Mean$ | $-0.05$ | 41.7 | $7.1 \times 10^{-5}$ |
| $Std$ | 2.018 | 26.9 | $2.87 \times 10^{-4}$ |
| $Min$ | $-3.5$ | 0 | $-3.23 \times 10^{-4}$ |
| $Max$ | 3.4 | 80.0 | $10.77 \times 10^{-4}$ |
| $25\%$ | $-1.8$ | 20.0 | $-1.11 \times 10^{-4}$ |
| $50\%$ | $-0.075$ | 42.5 | $2.00 \times 10^{-6}$ |
| $75\%$ | 1.7 | 65 | $1.33 \times 10^{-4}$ |

In table. 1, most of the essential statistics have been summed up. The dataset consists of 828 samples and three attributes; the first two attributes are the predictors and the 3$^{rd}$ one – which is current – is the response. Due to the vast differences between the first two features, the interquartile range (IQR) method has been chosen to scale the features before using them.

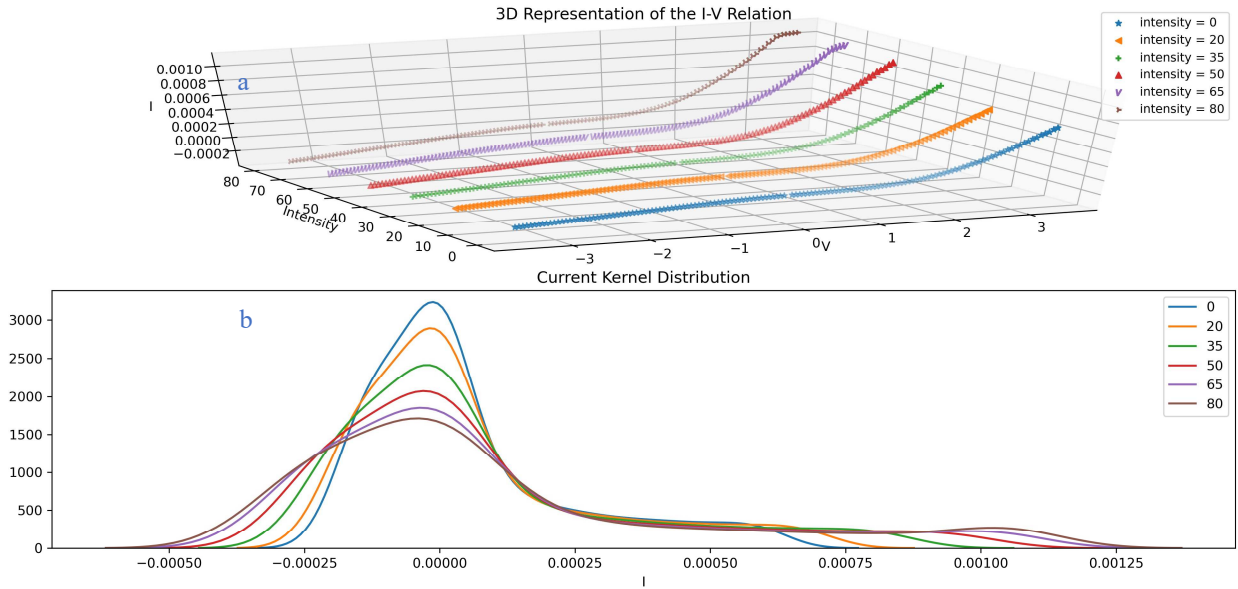

**Fig**. 2: a) 3D representation for I-V characteristic curve of NTCDA/p-Si photodiode under different illumination intensities, b) the kernel density distribution of the current based on different illumination intensities.

Based on Fig. 2, one can assume that despite the strong correlation between the voltage and the resulted current, the illumination intensity does affect the current distribution and hence there is a nonlinear relationship between the current and the illumination. Fig. 2.b represents the kernel distribution of the obtained values of the current at different illuminating intensities and confirms the nonlinear relation between them. Moreover, upon increasing the illumination intensity, the distribution function becomes broader. The obvious left-skewness is a typical asset of the organic/inorganic heterojunction devices. The absence of the sudden change in the positive side of the current distribution and smooth distribution at 50 $mW/cm^2$ confirms that the fabricated device shows its optimal performance at this illuminating intensity.

### 3.2.2 Model selection

Regarding the model selection step, different types of regression models have been examined and they are divided into two main groups; parametric models such as linear [43] and polynomial models [44] and nonparametric models such as KNN, Random Forest [45] and Support Vector Regression (SVR) [46], except for linear and polynomial kernels.

Herein, the parametric models are not a proper choice in the current regression problem, where the linear regression model is expressed in the closed-form solution as:

$$\omega = \left(X^T X\right)^{-1} X^T Y, \qquad (1)$$

where $X$ is an $m \times n$ matrix, and $Y$ is a vector of size $m \times 1$. From Eq. (1), it is evident that the linear model assumes a linear relationship between the dependent and the independent variables, and it assumes a homoscedastic pattern in the response variable. These two properties alone clarify why a linear model is not a proper choice for modelling the diode's behaviour.

On the other hand, nonparametric models do not assume a pre-determined form of the independent and the dependent variables. Instead, it learns the structure of the data itself. Moreover, a parametric model is described by a finite set of parameters, unlike other nonparametric models whose complexity depends mainly on the feature space of the dataset. Hence, according to the nonlinearity and the limited size of our dataset, the nonparametric regression models are more convenient.

A KNN model is chosen to be the fitting model because it does not assume an explicit form for the training data. Moreover, the testing time for such a model is $O(NK + Nd)$ [47] where $N$ is the number of samples (828), $d$ is the number of features (voltage and illumination intensity), and $K$ is the number of the closest points to the query testing sample. This means that the testing time is very fast on a typical CPU backend. The algorithm uses the inverse distance weighted average of the KNNs. There are different metrics, but the most general one is the Minkowski distance metric which is defined by the following equation: $(\sum_{i=1}^{n}|x_i - y_i|^p)^{1/p}$ where $x, y \in \mathbb{R}^2$. Furthermore, depending on the $p$ value, one can choose between the Euclidean, Manhattan, and Chebyshev metrics.

For hyper-parameter tuning, an exhaustive grid search technique was chosen for selecting the best $K$ value, the distance metric and its corresponding $p$ value, and the underlying algorithm that is used for calculating the distances such as K-D Tree, Ball Tree, and Brute Force, based on a cross-validation score of 13-folds. The fitness function for the grid search is the mean squared error metric. The $R^2$ score is also evaluated to check for the goodness of the fit [48].

The earlier approach, as will be discussed in the results section, achieves an excellent job in fitting the *I-V* data and reached an acceptable minimum error rate. However, our hopeful aim to obtain a lower error rate rather than that resulted from the KNN is still needed. Hence, another approach was adopted.

### 3.2.3 Artificial Neural Network (ANN)

The second modelling technique in the classical approach is ANN. ANN consists of small units called neurons which act as the neurons inside our brains. As shown in Fig. 3, one can formulate an equation that can describe it as follows:

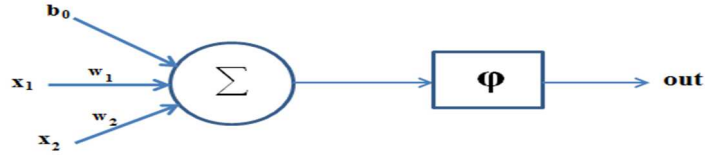

**Fig**. 3: *The working mechanism of a single neuron "perceptron".*

$$y = \phi(w \cdot x + b), \qquad (2)$$

where *y* is the output, *w* is the weight, *x* is a training sample, *b* is the bias, and $\phi$ is the activation function [49]. According to the universality theory [50,51], using connected layers made of neurons enables us to approximate any known function and fit its data. An example of such a network is the multilayer perceptron or a feedforward neural network as depicted in Fig. 4 which consists of an input layer, an output layer, and more than one hidden layer.

The size of the input and the output layers depends on the problem definition which – in our case – is two neurons and one neuron, respectively. Determining the size of the hidden layers and the neurons in each one of them is still an open debate [52,53]. Hence, we empirically chose those numbers; starting from a shallow network with a small number of units towards the current shape depicted in Fig. 4.

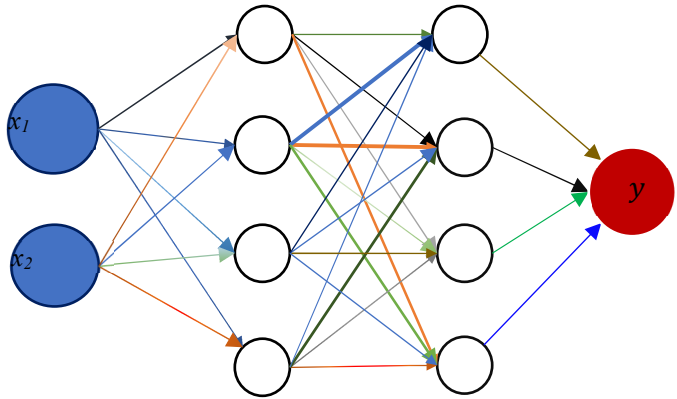

**Fig**. 4: Feedforward neural network architecture, where the faint colours indicate extremely low weights. The number of hidden layers in this model is 4. Further details can be found in the repository.

A feedback mechanism which is called backpropagation [54] is used to update the weights of each layer depending on the optimiser updating rule. The output from a neuron is just its activation's output, i.e. for an input $x_n$ where *n* is the number of that layer there would be an output vector $y_n$ which would be the input of $n + 1$ layer as follows:

$$\left. \begin{array}{l} y_n = \phi(x_n) \\ x_{n+1} = w_n \cdot y_n \end{array} \right\} \qquad (3)$$

where $\phi$ is the nonlinear activation function, and $w$ is the weight matrix between $n$ and $n+1$ layer. Setting $n=1$ to be the first layer and $N$ to be the last layer, we can then move to the next step which is backpropagation. Let us first define a cost function $E$:

$$E(X,W) = \frac{1}{m}\sum_{i=0}^{m}(y_i - w_i \cdot x_i)^2, \tag{4}$$

the error gradient at the last layer is:

$$\delta_N = \frac{\partial E}{\partial x_N}, \tag{5}$$

and the error gradient at the $n^{th}$ layer is:

$$\begin{aligned}\partial_n &= \frac{\partial E}{\partial x_n} \\ &= \frac{\partial E}{\partial x_{n+1}}\frac{\partial x_{n+1}}{\partial x_n}\end{aligned} \tag{6}$$

using Eq. (6), we can formulate a relation that describes how it is possible to propagate the errors backwards through the layers for each neuron as follows:

$$\begin{aligned}\delta_n &= \delta_{n+1}\frac{\partial w_n y_n}{\delta y_n}\frac{\partial \phi(x_n)}{\partial x_n} \\ &= \delta_{n+1} w_n \phi'(x_n).\end{aligned} \tag{7}$$

Intuitively, the error at the last layer is just the cost function multiplied by the derivative of the activation function. Finally, the third step takes care of updating the weights matrix according to the derivative between the cost function and the weights. This procedure is repeated until convergence occurs. The updates of the weight matrix are calculated using the Adam optimiser [55]. The activation function for all the layers – except the last one – is the RELU function [49]. The output layer's activation function is linear. The default parameters of the Adam's algorithm are left unchanged. Features are usually normalised between 1 and −1 before feeding them to the network. However, we have scaled them according to their mean and standard deviation, and that helped with making the convergence time a little shorter.

To overcome the difficulties of the earlier two approaches; data pre-processing and choosing the proper model, a more automated approach can help to decrease the time of data inspection, feature pre-processing and model selection. Hence, an AML technique is adopted [37].

### 3.2.4 Automated machine learning using Genetic programming

An automated approach is used to find the best modelling pipeline for the fabricated device. Tree-based Pipeline Optimisation Tool (TPOT) is an AML library that depends on (GP) [56] which is a well-known technique in Evolutionary Algorithms [57]. TPOT does the tedious job of pre-processing the data, feature engineering, model selection, hyper-parameter tuning, validation and finally testing. The idea lies in creating multiple copies of the dataset and feeding it to different pipelines in a tree-fashion way. It follows a simple procedure as follows:
 i. The initialisation of the population, randomly, to be the first generation.
 ii. Evaluation of the cost – fitness – function to check for the best individuals.
 iii. Repeating the same procedure again but with some restriction parameters such as time, mutation, crossover, and selection criteria to dump the weakest individuals and breed a new generation "the offspring."

The problem behind this approach is that in each run a new pipeline can be produced. That is the reason why we had it run several times, and the only thing we needed is to test each one to see if it is a good fit for our case. The chosen pipeline of this approach performs very well in terms of extrapolating unseen data and has an acceptable error rate. Moreover, it used the same regressor method that has been chosen for the first method. The big drawback was that it produced overfitted pipelines so that we had to pay careful attention to the value of the number of cross-validations and the population size. Further details are present in the results section.

The resulted pipeline as shown in Fig. 5 has three stages; the first one is modifying the data by copying it and adding a new predicted value by the XGBoost regressor [58], the second one is a feature union between the first stage and the data itself so that we have five features instead of two, the third stage is the estimator itself which is a KNN for the final prediction.

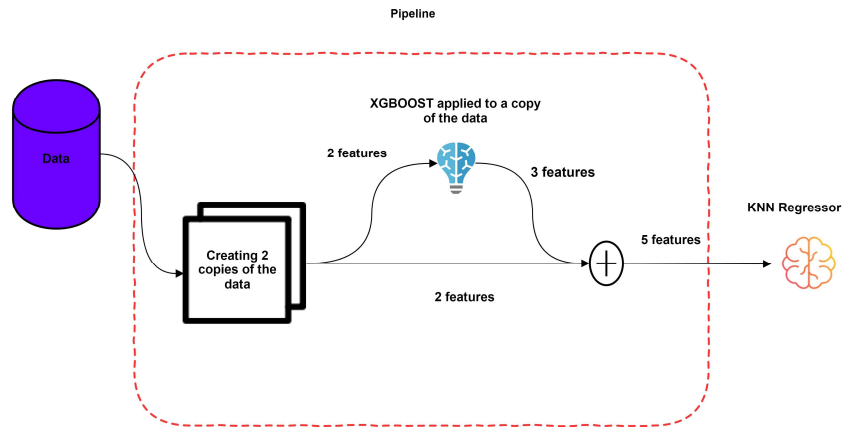

**Fig**. 5: Resulted pipeline of the TPOT method.

### 3.2.5 Quantum neural network (QNN)

Despite the considerable advancements in quantum computing technologies and the exciting and fruitful race between giant tech companies, quantum computers are still noisy and cannot operate at room temperatures. That is why we are living in the noisy intermediate-scale quantum computers (NISQ) era [59]. ML helps us to take full advantage of these devices despite their imperfect behaviour. Using a hybrid quantum-classical computational model [60,61] paves the way for using the classical part to do the repeated procedures that require memory and error-correcting subroutines whereas the quantum part performs the complex intractable classical operations.

Different software libraries provide this framework of computation. Pennylane [41] is an example of these libraries. It presents the concept of a quantum node (Qnode) where you can perform an operation by providing the inputs to the parameterised gates and then measure the desired observable. In this manner, the Qnode is not interrupted until we perform the measurement. This concept of hybrid computation makes it easier to realise a fully QNN where the backpropagation step is computed on the classical part and the feedforward step is computed on the quantum computer. Pennylane also provides a means to calculate the gradients of the quantum gates without the need for any classical interference [62].

Relating the aforementioned idea to the typical ML pipeline, we need to determine the way of feeding the classical data into the quantum device and to choose a proper model for the task. There are two main techniques in quantum information science to process the data; the Qubit model and the CV model [63–66]. Since our data are continuous, the CV model is chosen.

In the CV model, the information can be represented by the phase-space formalism [67,68], instead of the wavefunction formalism [69]. The former method is used as it's more convenient for our case. Generally, in this context, the data is encoded in terms of the $\hat{x}$ and $\hat{p}$ quadratures which satisfy the canonical relations by setting $\hbar = 2$:

$$[\hat{x}, \hat{p}] = 2i, \tag{8}$$

$$[\hat{a}, \hat{a}^\dagger] = I. \tag{9}$$

These operators can be written in terms of each other as follows:

$$\hat{x} = (\hat{a} + \hat{a}^\dagger),$$
$$\hat{p} = i(\hat{a}^\dagger - \hat{a}), \tag{10}$$

where $\hat{a}$ and $\hat{a}^\dagger$ are the annihilation and the creation operators, respectively. The basic state that can be converted into any desired state is the Vacuum state $|0\rangle$. The CV model has a set of gates that can be used to perform any mathematical operation or prepare any state. Moreover, due to its nature of processing the information, nonlinear gates are naturally present and are used to express its universality [70,71].

The single Qumode gates are called the Displacement gate $D(\alpha) = e^{(\alpha \hat{a}^\dagger - \alpha^* \hat{a})}$ where $\alpha \in \mathbb{C}$, the Squeezing gate $S(z) = e^{1/2(z^* a^2 - z a^{\dagger 2})}$ where $z = re^{i\theta}$, and the Rotation gate $R(\varphi) = e^{i\varphi \hat{a}^\dagger \hat{a}}$ where $\varphi \in [0, 2\pi]$. The nonlinear gates are the Kerr gate $K(\hat{k}) = e^{ik(\hat{a}^\dagger \hat{a})^2}$ where $k \in \mathbb{R}$ and the Cubic Phase gate $V(\gamma) = e^{\left(i\frac{\gamma}{6}(\hat{a}+\hat{a}^\dagger)^3\right)}$ where $\gamma \in \mathbb{R}$ [63,64,68,72]. The nonlinear gates help with constructing higher-order Hamiltonians which are an incredible feat over the Qubit model, and one does not require more than a single variable nonlinear gate for this construction.

The architecture of the QNN is adapted from [32]. As depicted in Fig. 6, we are only using one Qumode with a repeated block of gates. Each block represents a layer in terms of classical neural networks. The classical data is embedded in the quantum device by preparing a Displaced Squeezed state $|\alpha, z\rangle$ where $\alpha, z \in \mathbb{C}$ [73]. It is created by applying a Squeezing gate then followed by a Displacement gate.

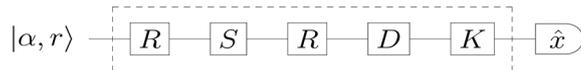

**Fig.** 6: The configuration of the quantum circuit used for modelling. The grouped gates represent a single layer of the QNN. The dashed box is repeated 8 times. The $r$ in the state preparation represents the magnitude of the squeezing gate.

Embedding classical data into quantum states is an active area of research [74,75]. Since we are only interested in measuring the $\hat{x}$ quadrature and we're simulating a classical NN within a quantum framework, only the absolute values of $\alpha$ and $z$ are being used to encode the classical data of the *I-V* curve. The voltage values are scaled between [-1.1, 1] to be embedded by the displacement gate. The illumination values are scaled between [0, 0.8] to be used as the squeezing magnitudes.

This scheme of data encoding is depicted in Fig. 7.a where the Wigner function has been plotted for a Displaced Squeezed state of the values $|-1.0, 0.8\rangle$ and the learned state of the QNN after applying the parametrised circuit as shown in Fig. 7.b. In terms of the quadratures, this would resemble a translation in the $\hat{x} \rightarrow S\hat{x} + d$ where $S$ is the Squeezing gate and its effect is simply narrowing the $\hat{x}$ quadrature by $e^{-r}$ and $r$ is the $\mathfrak{Re}(z)$, whereas $d$ is the $\mathfrak{Re}(\alpha)$ and it translates the $\hat{x}$ from the origin by its magnitude. On the other hand, the $\hat{p}$ quadrature is only affected by the $S$ gate as it is stretching by $e^r$.

According to [76], this mapping is critical to machine learning since it can be used as a kernel for feature mapping as the distance between two Displaced Squeezed states is calculated explicitly as follows [77]:

$$d(\langle \alpha_1, z_1 | \alpha_2, z_2 \rangle) = \sqrt{2\left(1 - |\langle \alpha_1, z_1 | \alpha_2, z_2 \rangle|^2\right)}, \tag{11}$$

where

$$\langle \alpha_1, z_1 | \alpha_2, z_2 \rangle = \left(\frac{(1-|z_2|^2)(1-|z_1|^2)}{(1-z_2 z_1^*)^2}\right)^{1/4}$$
$$\times \exp\left\{-\left[2(1-z_2 z_1^*)\right]^{-1}\left((\alpha_2 + z_2\alpha_2^*)(\alpha_2^* + z_1^*\alpha_2)\right.\right. \tag{12}$$
$$\left.\left. -2(\alpha_2 + z_2\alpha_2^*)(\alpha_1^* + z_1^*\alpha_1) + (\alpha_1 + z_2\alpha_1^*)(\alpha_1^* + z_1^*\alpha_1)\right)\right\}.$$

When the illumination is zero, then the prepared state is simply a Coherent state which acts as a radial basis kernel. This mapping procedure serves the regression problem we are trying to solve as we are only using a single Qumode to follow the architecture of [32].

Following the same footsteps of classical NNs, this leaves us with determining the number of layers, the learning rate, the size of each batch and weight initialisation. The parametrised circuit consists of 8 layers; each layer consists of 5 parameters, i.e. it only has 40 parameters to be updated by the Adam optimiser. The batch size is 32, and we tried different settings for the learning rate starting from 0.001 to 0.01. The weight initialisation was drawn from a random sample distribution with a standard deviation ranging from $[10^{-4}, 10^{-2}]$. The cut-off range was set to 18 for the Fock space.

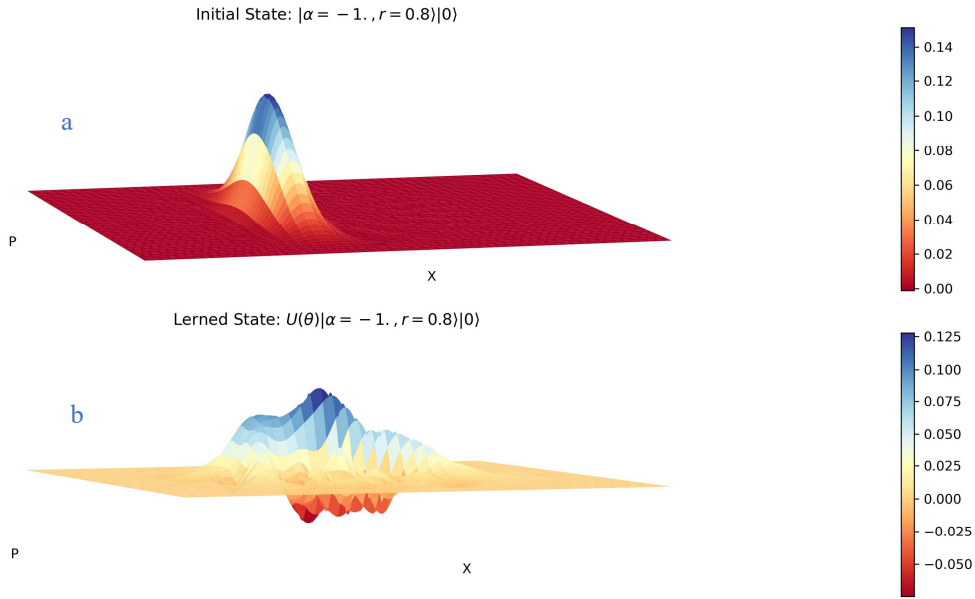

**Fig. 7**: a) is the Displaced Squeezed state. We can see clearly that the state is shifted to the negative side of the $\hat{x}$ quadrature and it's stretched in the $\hat{p}$ direction. On the other hand, b) represents the final state of the QNN with negative values in Wigner function which indicates the effect of the non-linear effect of the Kerr gate.

# 4 Results and Discussion

## 4.1 Optical Characteristics of NTCDA Thin Film

For inspecting the most suitable spectral range of detection, the optical absorbance of NTCDA thin film is measured in the range $190 - 2500\ nm$. The absorption profile shown in Fig. 8.a confirms the high UV absorption of NTCDA extended to the visible range with very low IR absorption. As observed from the inset of Fig. 8.a, the intensity of the UV absorption peaks is more significant than that of the visible region which suggests a good selectivity to detect even a weak UV signal in a robust visible light background.

The featured absorption peaks exhibited in the NTCDA absorption spectra implies the massive generation of photoexcited charge carriers in the NTCDA thin film. This leads to a high photocurrent and a high photoresponse photodetector. The optical band gap energy is estimated using absorption spectrum fitting (ASF) procedures [78,79]. The value of the indirect energy gap can be estimated by extrapolating the linear portion of the plot of $(Abs.\lambda^{-1})^{1/2}$ versus $\lambda^{-1}$ to $[(Abs.\lambda^{-1})^{1/2} = 0]$ as depicted in Fig. 8.b which is corresponding to $1/\lambda_g$, where $\lambda_g$ is the wavelength related to the energy gap. The optical band energy can be estimated by $[E_g = hc/\lambda_g]$. The estimated values for the onset energy gap and the optical energy gap are about 2.84, and $3.75\ eV$, respectively.

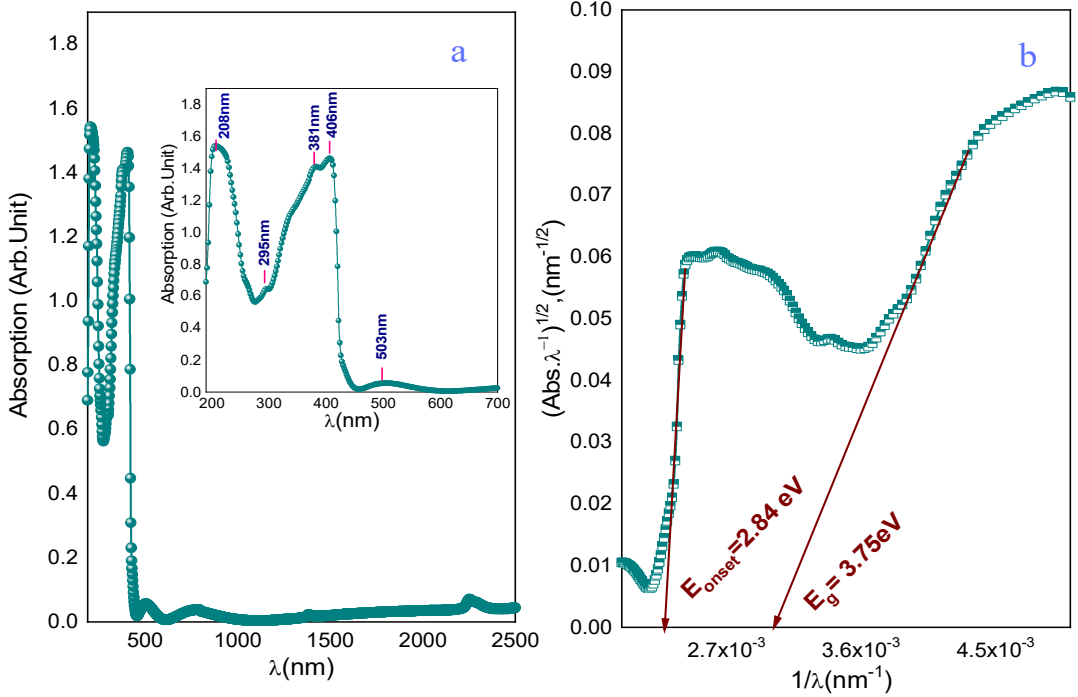

**Fig. 8:** a) Absorbance profile of NTCDA thin film, while the inset shows zoom-in on UV-visible absorbance and b) estimating the onset and indirect energy gap of NTCDA thin film using ASF method.

## 4.2 Electrical Characteristics of Au/NTCDA/p-Si/Al Photodiode

Fig. 9 depicts the semi-logarithmic current-voltage relation of Au/NTCDA/p-Si/ Al photodiode under the dark and illumination conditions at room temperature. A distinct rectification behaviour is obtained with ~3 and ~ 4 for dark and 80 $mW/cm^2$ irradiance at ± 3.5 V, respectively. A monotonic increase in the produced photocurrent is accompanied by the increase of light intensity. This increase emphasises the formation of neutral *Frenkel* excitons which dissociate generating free electrons and holes which are swept under the influence of the biasing voltage [38,80].

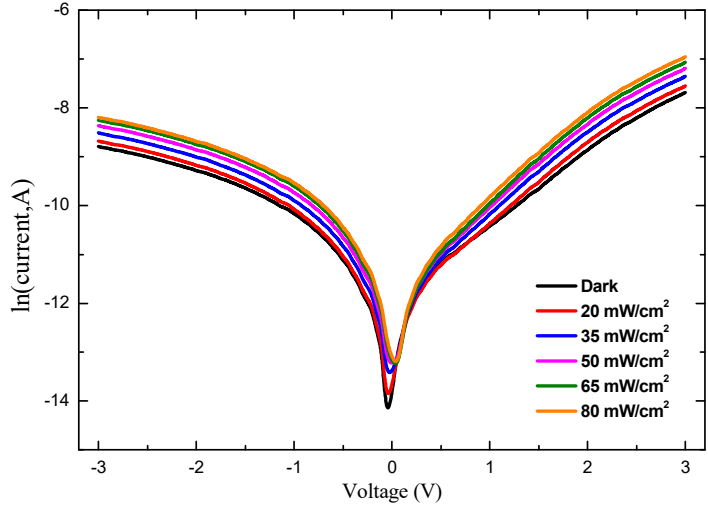

**Fig. 9:** Semi-logarithmic plot of *I-V* characteristics of NTCDA/p-Si photodiode under different illumination intensities.

The thermionic emission theory was employed to fit the linear region of the forward-biased *I-V* curve at low voltages ($V_F < 0.15\ V$) using the following relations [38,81]:

$$I = I_0 \exp\left(\frac{q(V - IR_s)}{nk_BT}\right)\left[1 - \exp\left(\frac{-q(V - IR_s)}{k_BT}\right)\right], \tag{13}$$

where *I* is the diode current, $I_0$ is the reverse saturation current, *n* is the ideality factor, *q* is the electron charge, $k_B$ is Boltzmann constant, *T* is the temperature, $R_s$ is the series resistance, and *V* is the applied voltage. The reverse saturation current and the barrier height at zero voltage, $\Phi_{Bo}$, are interrelated by the following equation [38,81]:

$$\phi_{B0} = \frac{k_BT}{q}\ln\left(\frac{AA^*T^2}{I_0}\right), \tag{14}$$

where *A* is the active area of the device and $A^*$ is the effective *Richardson* constant.

The ideality factor, *n*, was estimated from the slope of the linear region of the *I-V* curve in forward biasing with neglecting the series resistance in the low voltage region $V_F < 0.18\ V$ using the following relation [38]:

$$n = \frac{q}{k_BT}\left[\frac{dV}{d(\ln I)}\right], \tag{15}$$

Upon illumination, the estimated values of the ideality factor were decreased from 4.98 (in the dark) to 3.78 (under 80 $mW/cm^2$ irradiance). The high values of ideality factor over-unity may be attributed to many factors such as the tunnelling process [82], the image-forces effect [83], series resistance [84], inhomogeneity of barrier height and the existence of interface states [85,86]. Herein, the series resistance and the interfacial layer existence have the primary responsibility of the high value of ideality factor. Moreover, a slight increase in the values of zero-bias barrier height from 0.7 $eV$ (in the dark) to 0.72 $eV$ (under 80 $mW/cm^2$ irradiance) which may be attributed to the passivation of Si surface by NTCDA [81].

Inherently, the series resistance has a crucial impact on the performance of any electronic device, where these types of devices always exhibit a high series resistance. In this essence, many models were developed to estimate the values of series resistance, but each model has its limitations [87]. Subsequently, the series resistance of the fabricated device was estimated using a modified *Nord's* model, which is applicable to the whole voltage range [86–89].

A significant decrement in the series resistance from 2.96 $k\Omega$ to 0.10 $k\Omega$ upon increasing the illumination intensity from 0 $mW/cm^2$ to 80 $mW/cm^2$ is observed. This decrement may be explained as follows; under the influence of illumination, the photo-generated charge carriers would be trapped at either at p-Si/NTCDA interface or NTCDA/ electrode interface. Increasing the illumination intensity increases the density of photo-generated charge carriers and decreases the probability of trapping charges, decreasing the series resistance [86,90].

The estimated high values of ideality factor indicate that thermionic emission is no longer suitable for interpreting the transport mechanism, so to investigate the operating charge transport mechanisms in Au/NTCDA/p-Si/Al photodiode, the double logarithmic relation between current and the voltage in forward biasing in dark and under 80 $mW/cm^2$ irradiance are plotted as shown in Fig. 10.a. Two distinct regions with different slopes are obtained; region (**I**) of slope equals one and region (**II**) of slope equals 2.62. The dominant transport mechanism in the region (**I**) is the Ohmic transport mechanism [91], while in the region (**II**) the dominant transport mechanism is the space charge limited conduction (SCLC) that is dependent on trap level density and interface states [91,92].

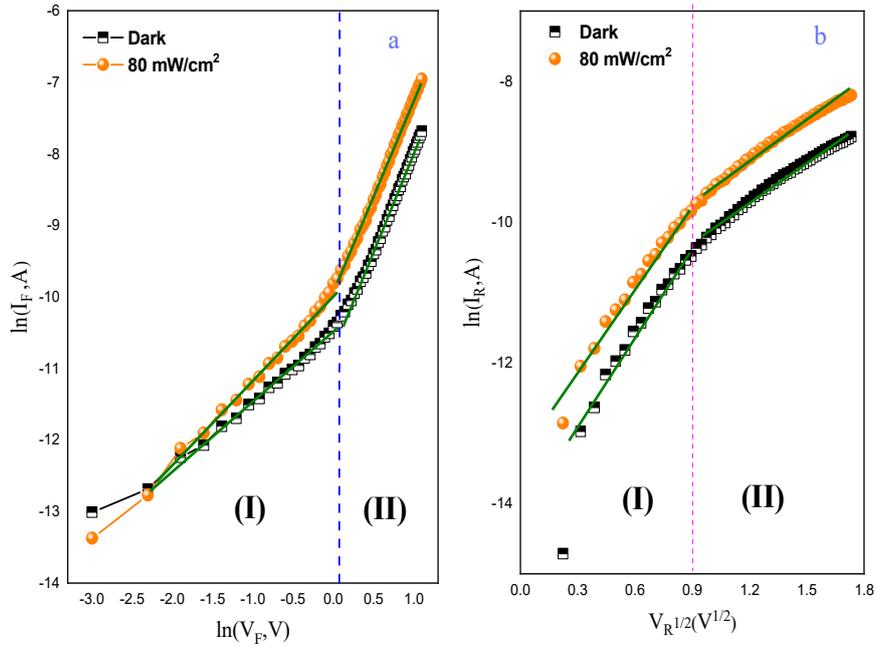

**Fig. 10**: a) $ln(I_F)$ versus $ln(V_F)$ and b) $ln(I_R)$ versus $V_R^{1/2}$ for NTCDA/p-Si photodiode in dark and under 80 $mW/cm^2$ illumination.

Furthermore, the transport conduction mechanism in reverse biasing of Au/NTCDA/p-Si/Al photodiode is investigated by plotting the relation between $V_{R^{1/2}} - \ln(IR)$ in the dark and under 80 $mW/cm^2$ as shown in Fig. 10.b. Two linear regions of different slopes were obtained which may be attributed to either *Schottky* or *Poole-Frenkel* mechanism.

According to *Schottky* conduction mechanism, the reverse current can be represented as following [92]:

$$I_{Sc,R} = AA^* T^2 \exp\left(\frac{-\phi_B}{k_B T}\right) \exp\left(\frac{\beta_{Sc} E^{1/2}}{k_B T}\right) \qquad (16)$$

where $E$ is the applied electric field, and $\beta_{S_c}$ is the *Schottky* field lowering coefficient. Based on *the Poole-Frenkel* emission mechanism, the reverse current can be represented as following [92]:

$$I_{PF,R} = I_S\,A\exp\left(\frac{\beta_{PF}E^{1/2}}{k_B T}\right), \qquad (17)$$

where $\beta_{PF}$ is the *Poole-Frenkel* coefficient. The value of $\beta$ is given by [92]:

$$\beta = \left(\frac{q^3}{b\pi\varepsilon\varepsilon_o}\right)^{1/2}, \qquad (18)$$

where $b$ is a constant equals 1 for the *Poole-Frenkel* mechanism and equals 4 for *Schottky* mechanism [92], $\varepsilon$ is the dielectric constant of the material ($\varepsilon = 2.89$ [93] ), $\varepsilon_o$ is the permittivity of free space ($8.85 \times 10^{-12}$ F/m$^2$) and $q$ is electron's charge. The calculated values of $\beta$ were about $4.46 \times 10^{-5}\ eVm^{0.5}V^{0.5}$ for *Poole-Frenkel* mechanism and $2.23 \times 10^{-5}\ eVm^{0.5}V^{0.5}$ for *Schottky* mechanism. The experimental values of $\beta$ estimated from the slope of the straight lines in the region (**I**) and region (**II**) were about $4.89 \times 10^{-5}eVm^{0.5}V^{0.5}$ and $2.05 \times 10^{-5}\ eVm^{0.5}V^{0.5}$, respectively. Hence, the transport mechanism in the low voltage region (**I**) is interpreted as *a Poole-Frenkel* mechanism, while in the high voltage region (**II**) is interpreted as a *Schottky* emission mechanism.

### 4.3  Optoelectronic Characteristics of Au/NTCDA/p-Si/Al Photodiode

An explicit investigation of photosensitivity of the fabricated photodiode was performed by studying the dependence of photocurrent density, $J_{Ph}$, on illumination intensity at different biasing voltages as shown in Fig. 11.a. The fabricated device exhibits a linear performance with the irradiance up to $65\ mW/cm^2$, but exceeding this limit makes the photodiode saturates. The linearity of photocurrent density and irradiance relation is more obvious in the high biasing voltages than lower voltages. Fig. 11.b shows the double logarithmic relation between the photocurrent, $I_{Ph}$, and incident power, $P$, where the estimated slopes ($\sim 1.4$) conclude that this architecture can be utilised in optoelectronic applications to detect low-intensity signals.

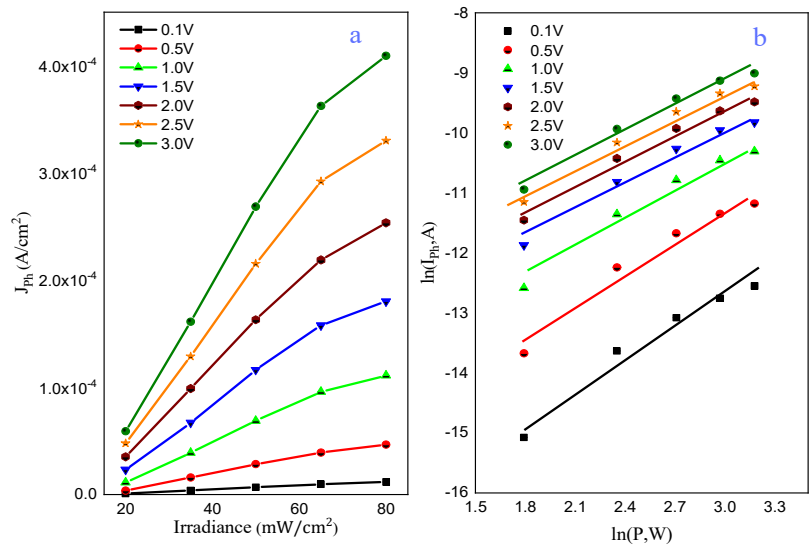

**Fig. 11:** a) The irradiance-photocurrent density relation and b) $lnP$ versus $lnI_{Ph}$ for NTCDA/p-Si photodiode at different biasing voltages.

The quantitative evaluation of the optoelectronic performance of the existing architecture necessitates an accurate estimation of several figures of merits:

$$R = \frac{J_{Ph}}{P}, \tag{19}$$

$$D^* = R \Big/ \sqrt{2qJ_{dark}} = D \times \sqrt{A}, \tag{20}$$

$$EQE = \frac{hc}{\lambda q} \times R \times 100, \tag{21}$$

$$I_N = NEP.R = \frac{R\sqrt{A}}{D^*}, \tag{22}$$

where *NEP* is the noise equivalent power. Fig. 12 shows the voltage dependence of responsivity, specific detectivity, external quantum efficiency and noise current at different illumination intensities. An apparent increase of responsivity, specific detectivity, and external quantum efficiency with increasing the illumination intensity up to $65\ mW/cm^2$, which means that this architecture operates for detecting a light signal of illumination lower than $65\ mW/cm^2$. The decrease of photoresponsivity of the fabricated device at intensity $80\ mW/cm^2$ may be attributed to the decrease of sufficient photo-induced trap states density at the interface [94].

The calculated specific detectivity which determines the resolution limit that the detector can distinguish from the noise [95] is in the range of 109 Jones as shown in Fig. 12.b which proves the sensitivity of the fabricated device [94]. From Fig. 12.c, the EQE of the fabricated photodiode records the highest value at $65\ mW/cm^2$ illumination and $-3.5\ V$ biasing voltage. Inherently, there are three contributions to noise current in photodetector that should be considered; thermal noise or Johnson noise current ($IJ$), shot noise current ($ISh$) and flicker noise current ($IFl$) [96].

The estimated values of noise current were about $0.4\ pA$ and $7\ pA$ at $0\ V$ and $-3\ V$, respectively. The absence of effective influence of the light intensity on the noise current, as shown in Fig. 12.d proves that the shot noise from dark current density is the main responsible for noises in the photodiode [96,97]. Table 2 shows a comparison study between the estimated parameters of the present device and earlier work.

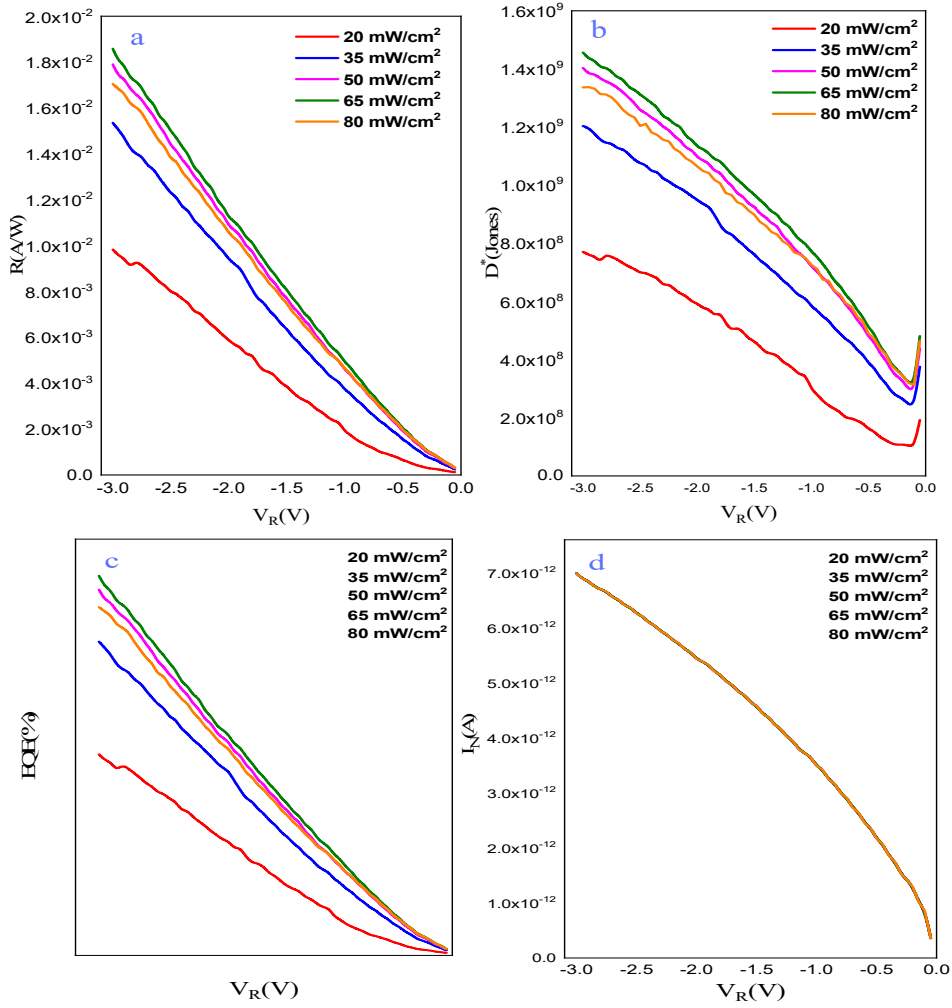

**Fig. 12:** The voltage dependence of a) responsivity, b) specific detectivity, c) external quantum efficiency and d) noise current of the fabricated device at different illumination intensities.

**Table 2:** The performance parameters of some UV and visible Si-based photodiodes.

| Photodiode structure | Detection band | Responsivity (mA/W) | Specific detectivity (Jones) | EQE (%) | Rise time | Fall time | Ref |
|---|---|---|---|---|---|---|---|
| NTCDA/p-Si at 65 $mW/cm^2$ | UV | 19 (at= -3 V) | $1.5 \times 10^9$ (at= -3 V) | 11.85 | 433 ms | 495 ms | Present work |
| Cu(acac)$_2$/n-Si at 80 $mW/cm^2$ | UV and Visible | 4.7 (at= -3.5 V) | $2.7 \times 10^9$ (at= -3.5 V) | 5.53 | 339 ms | 1046 ms | [86] |
| TiOPc/p-Si at 180 $mW/cm^2$ | IR | $1.63 \times 10^{-2}$ (at= -2V) | $1.22 \times 10^7$ (at= -2 V) | 0.25 | 23 µs | 34 µs | [98] |
| PTCDI/p-Si at 200 $mW/cm^2$ | IR | 0.2 (at= 0 V) | $7 \times 10^7$ (at= 0 V) | – | 300 ns | 300 ns | [99] |
| P-quaterphenyl /p-Si | UV and Visible | 9 (at= -1 V) | $1.14 \times 10^8$ (at= -1 V) | – | 25 µs | – | [100] |
| Au/α-6T/n-Si at 2560 $mW/cm^2$ | UV and Visible | 0.2 (at = -2.5 V) | $1.2 \times 10^8$ (at= -2.5 V) | 0.3 | 410 ms | 400 ms | [101] |

The photoresponse speed of any optical device plays a crucial role in evaluating its optoelectronic performance, where almost all the optical devices require picking up a light signal with a certain bandwidth with fast response. The photoresponse speed which measures the quickness of extracting the photo-generated charge carriers to the external electrical circuit is dependent on many factors such as active layer thickness, the choice of interfacial layer material and applied biasing voltage [102].

Fig. 13 shows the transient photocurrent response of the fabricated photodiode under the influence of the square UV signal of 0.1 Hz frequency at different illumination intensities and 1V biasing voltage. The illustrated time-tracked photocurrent is evidence of the strong and fast photoresponse of the existing architecture with good repetition and stability. Furthermore, it is observed that Au/NTCDA/p-Si/Al photodiode has a residual photo-current in the OFF state which increases slightly with increasing the illumination intensity.

This may be explained in terms of enhancement of NTCDA photoconductivity due to photo-thermal energy emitted of the incident light [86]. Fig. 14(a-d) reveals the fall time and rise time assessment for the fabricated photodiode under 20, 35, 65 and 80 $mW/cm^2$ illuminating intensity. A clear decrease in fall and rise times is noticed upon increasing the illumination intensity. Moreover, the ON/OFF ratio for different irradiance values at different reverse biasing voltages is estimated as shown in Fig. 14.e. For high irradiance values (> 20 $mW/cm^2$), the most appropriate biasing voltage is $1V$ to get a higher ON/OFF ratio, while at 20 $mW/cm^2$ increasing the applied voltage increases the ON/OFF ratio.

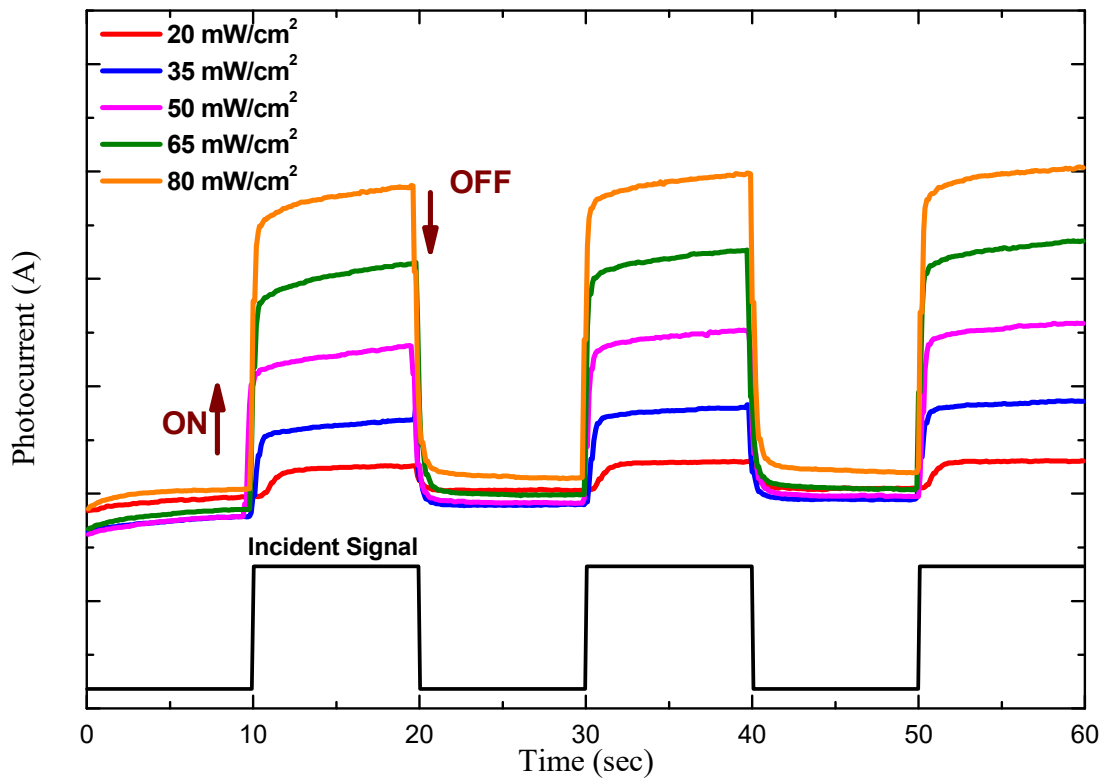

**Fig.13:** Repeatable ON/OFF switching behaviour of the fabricated photodiode under different illumination intensities at 1 $V$.

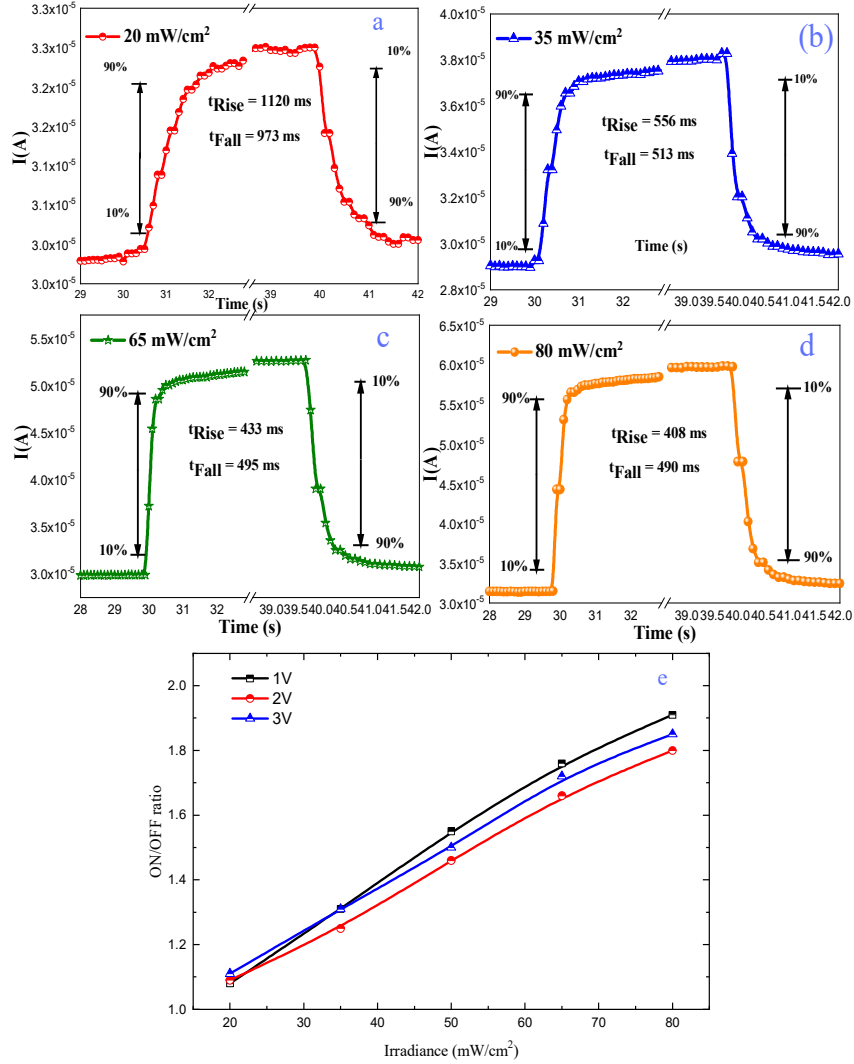

**Fig.14:** The analysis of the rise and the fall time of Au/NTCDA/p-Si/Al photodiode under **a)** 20 $mW/cm^2$, **b)** 35 $mW/cm^2$, **c)** 65 $mW/cm^2$, **d)** 80 $mW/cm^2$ irradiance and **e)** ON/OFF ratio – irradiance relation at different biasing voltages.

## 4.4   Modelling the Optoelectronic Characteristics of Au/NTCDA/p-Si/Al Photodiode

To make our results reproducible and consistent, we paid careful attention to the randomness that may occur while splitting the data into training and testing sets or even the stochastic nature of the NNs. Moreover, we trained all the algorithms on the same partitioning so that we can compare the performance of each one with the others. We set 15% of the data to be the testing set and the other 85% to be the training data. The traditional ML algorithms were adopted from Scikit-Learn and TPOT. For the ANN part, Keras was used. Pennylane and StrawberryFields were used to model the QNN and to study its characteristics. The results of the four models are presented in table 3.

In Fig. 15, the connected coloured lines represent the prediction of the trained model and the scattered points are the target values. Fig. 15.a depicts the predictions of the KNN model. It achieved an MSE of $2.67 \times 10^{-11}$ and the training error is 0. The 0 error is typical behaviour from such a model as the training procedure simply memorises the data. The IQR pre-processing step was the best way for preparing the data for the KNN. The model's parameters are $K = 4$, $p = 4$. The metric function is the Minkowski, and the searching algorithm is the Brute Force.

Despite being fed with raw data directly, the predicted curves of the TPOT in Fig 15.b shows very similar behaviour to the KNN model since the last step of the pipeline is also a KNN regressor. Even though the validation error is lower than the one in the KNN case, the testing error is higher in the TPOT model. That arises from the fact that the TPOT training procedure tends to overfit the training data. To avoid such a scenario, we had to configure its dictionary of parameters to remove specific methods of feature engineering such as Radial Basis Function Sampler and Polynomial Features. We also changed the generation parameter to be 50 and the crossover to be 0.15.

Fig. 15.c depicts the prediction of the ANN model. Due to the stochastic nature of the NN training, the loss kept oscillating but with a minimal variance after 10000 epochs as depicted in Fig. 16.a. With a batch size of 32, a learning rate of $10^{-3}$, and 6 layers, the ANN achieved the lowest error rate on the testing set $1.0897 \times 10^{-11}$. The ANN is the most complex model with approximately $10^3$ optimised parameters. This approach alongside the earlier two models can efficiently model the photodiode's behaviour. Their results were tested analytically, and they produced the same figures in sections 4.1, 4.2, and 4.3.

In Fig 15.d, the QNN predicted curves are plotted and in Fig. 16.b the training and testing losses are depicted. The QNN model didn't produce the best results since the target values are extremely small, and we only used the values that affect the $\hat{x}$-quadrature only. However, with a single Qumode and only 40 parameters, we managed to simulate what the ANN has achieved. Moreover, the loss function only took less than 500 epochs to reach the lowest error rate for both testing and training errors with a smooth curve. This means a higher convergence rate than the classical ANN. In addition to that, the QNN didn't overfit the training data and managed to capture the real trend of the *I-V* curve by predicting extremely small values when the input data were zero for the Displacement parameter.

We had to scale the target values by a factor of $10^3$ because of two reasons. The first one is that the difference between the expectation values is very small for the prepared quantum states that have a small input voltage. The second reason is that the optimiser tends to update the active set of parameters; the Displacement and the Squeezing gates' parameters, with high values so that the expectation value would be small enough, and the loss function would become smaller. To prevent the optimiser from increasing the active parameters we added a regularisation term to the loss function of Eq. (4) as follows:

$$E = \frac{1}{m}\sum_{i=0}^{m}(y_i - \langle\psi_i|\hat{x}|\psi_i\rangle)^2 + \lambda\frac{1}{m}\sum_{i=0}^{m}\left(1-\left|\langle\psi_i|\mathrm{I}|\psi_i\rangle\right|\right), \qquad (23)$$

where $y$ is the target value and $|\psi\rangle$ is $U(\theta)|\alpha,z\rangle$ i.e. after applying the parameterised quantum circuit to the initialised quantum state and $\theta$ represents the variational parameters that are being optimised. The second term consists of $\lambda$ which has been chosen to be 0.01 and the rest is the trace of the quantum system for each sample. If the trace is equal to 1, the loss function reduces to the form of Eq. (4). This added term preserves the sanity of the quantum state within the truncated Fock space and keeps it normalised. We also avoided using the Cubic Phase gate as it made us increase the truncation space to be beyond 20 for a healthy quantum system. However, the convergence rate is faster using the cubic phase gate.

Table 3: The loss values for each model and the relative usability of each method according to the required knowledge for each one to be developed properly.

| Model | Train | Test | Validation | Usability |
|---|---|---|---|---|
| KNN | 0 | $2.67 \times 10^{-11}$ | $6.07 \times 10^{-11}$ | Medium |
| TPOT | 0 | $3.58 \times 10^{-11}$ | $5.38 \times 10^{-11}$ | Easy |
| ANN | $2.382 \times 10^{-11}$ | $1.0897 \times 10^{-11}$ | - | Hard |
| QNN | $1.2655 \times 10^{-10}$ | $9.695 \times 10^{-11}$ | - | Very Hard |

In terms of practical software developments, the TPOT method is the easiest one to be used as it only requires having a basic knowledge of software programming and a solid understanding of different fitting metrics and criteria. The rest is completely automated by its internal API. The KNN model requires a solid understanding of ML basics and how to prepare the data for such a distance-based algorithm. The last two algorithms require an expert level especially the QNN model. In terms of pure fitting criteria, we recommend the KNN model – with a higher number of $k$-folds for hyperparameter tuning – to be the go-to model for such situations.

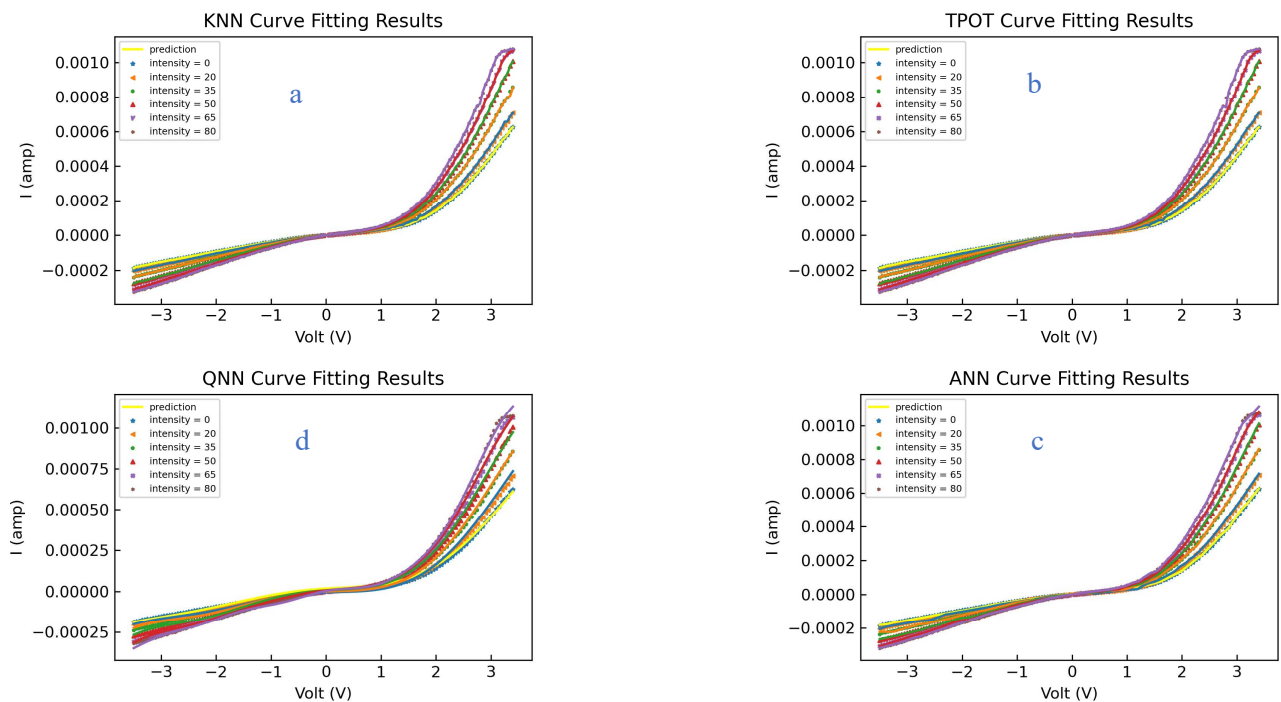

**Fig**. 15: The fitted experimental data of I-V characteristic curve of NTCDA/p-Si using a) KNN, b) TPOT, c) ANN and (d) QNN models.

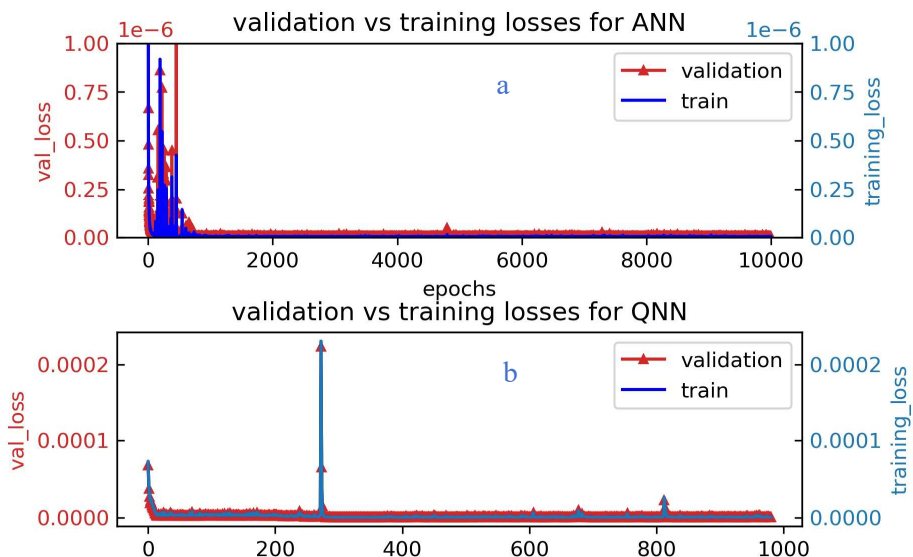

**Fig**. 16: a) is the training and validation loss of the ANN, and b) is the same for the QNN.

## 5 Conclusion

Organic/inorganic heterojunction based on NTCDA/p-Si is successfully fabricated for UV light detection manner. According to the obtained values of spectral responsivity, linear-dynamic range, specific detectivity, signal to noise ratio, and response time for the fabricated photodiode under the influence of UV light of intensity $20 -$ 80 $mW/cm^2$, the current device is suggested for many optoelectronic applications. These obtained results ensure that the optimum light intensity – where this detector works efficiently – is 50 $mW/cm^2$.

Regarding the modelling section of this research, adopting a regression analysis approach resulted in three accurate classical algorithms that can be used to model and predict the behaviour of the fabricated device. These models can both extrapolate and interpolate unknown current values under different illuminations and voltage settings. Furthermore, they can predict extreme values near zero voltage with a minimal error rate. This can help other researchers conduct their studies without going through the costly manufacturing process.

Using the fourth model, i.e. the QNN, we managed experimentally to show the superiority of the Qumode as it can encode floating-point data easily. In addition to that, a single Qumode – with only 40 parameters – has the capability of reproducing what a classical fully connected NN can do with thousands of parameters. This is a powerful indicator of the universality of the CV-QNN model.

In our future work, we will explore the possibility to combine the QNN and the ANN into a single computational paradigm. In addition to that, we will study the effect of data embedding on the result. It will be an interesting direction if we study the characteristics of the material itself using ML and quantum computing to explore much more possibilities and discover new materials.

### Acknowledgements

The authors would like to express their most profound appreciation to Prof. Zoltán Zimborás from Wigner Research Centre for Physics, Hungarian Academy of Sciences and Dr Josh Izaac from the University of Western Australia for their helpful discussions and keynotes.

### Funding

This research did not receive any specific grant from funding agencies in the public, commercial, or not-for-profit sectors.

### Conflicts of interest/Competing interests

The authors have no affiliation with any organization with a direct or indirect financial interest in the subject matter discussed in the manuscript.

### Code & Data Availability

The code and the dataset can be found in this [GitHub](GitHub) repository.